\documentclass[lettersize,journal]{IEEEtran}
\usepackage{amsmath,amsfonts}
\usepackage{algorithmic}
\usepackage{array}
\usepackage[caption=false,font=normalsize,labelfont=sf,textfont=sf]{subfig}
\usepackage{textcomp}
\usepackage{stfloats}
\usepackage{url}
\usepackage{verbatim}
\usepackage{graphicx}
\def\BibTeX{{\rm B\kern-.05em{\sc i\kern-.025em b}\kern-.08em
    T\kern-.1667em\lower.7ex\hbox{E}\kern-.125emX}}
\usepackage{balance}

\newcommand{\ie}{\textit{i}.\textit{e}.}
\newcommand{\eg}{\textit{e}.\textit{g}.}
\newcommand{\etc}{\textit{etc}}
\usepackage[mathscr]{euscript}
\usepackage{multirow}
\usepackage{color}

\usepackage{booktabs} 
\newcommand{\dt}[1]{\fontsize{8pt}{0.1em}\selectfont (#1)}
\usepackage{fancyhdr}
\usepackage{hyperref}

\begin{document}
\title{SRRT: Exploring Search Region Regulation for Visual Object Tracking}
\author{Jiawen Zhu, Xin Chen, Pengyu Zhang, Xinying Wang,
	Dong Wang,  Wenda Zhao, Huchuan Lu

\thanks{
This work was supported in part by the National Natural Science Foundation of China under Grant No. 62293540, Grant No. 62293542 and Grant No. U23A20384; and in part by the Talent Fund of Liaoning Province (no. XLYC2203014).

Jiawen Zhu, Xin Chen, Pengyu Zhang, Dong Wang, Wenda Zhao, and Huchuan Lu are with the
School of Information and Communication Engineering, Dalian University
of Technology, Dalian 116024, China  (e-mail:
jiawen@mail.dlut.edu.cn; wdice@dlut.edu.cn; lhchuan@dlut.edu.cn). 

Xinying Wang is with the School of
Computer Science and Technology, Dalian University
of Technology, Dalian 116024, China (e-mail: wangxinying@mail.dlut.edu.cn). 
}
}

\markboth{JOURNAL OF \LaTeX~  CLASS FILES, VOL. 14, NO. 8, AUGUST 2021}%
{SRRT: Exploring Search Region Regulation for Visual Object Tracking}

\maketitle

\begin{abstract}
	The dominant trackers generate a fixed-size rectangular region based on the previous prediction or initial bounding box as the model input, \ie,~search region. 
	While this manner obtains promising tracking efficiency, a fixed-size search region lacks flexibility and is likely to fail in some cases, \eg, fast motion and distractor interference. 
	Trackers tend to lose the target object due to the limited search region or experience interference from distractors due to the excessive search region.
	Drawing inspiration from the pattern humans track an object, we propose a novel tracking paradigm, called Search Region Regulation Tracking (SRRT) that applies a small eyereach when the target is captured and zooms out the search field when the target is about to be lost.
	SRRT applies a proposed search region regulator to estimate an optimal search region dynamically for each frame, by which the tracker can flexibly respond to transient changes in the location of object occurrences.
	To adapt the object's appearance variation during online tracking, we further propose a locking-state determined updating strategy for reference frame updating. 
	The proposed SRRT is concise without bells and whistles, 
	yet achieves evident improvements and competitive results with other state-of-the-art trackers on eight benchmarks. 
	On the large-scale LaSOT benchmark, SRRT improves SiamRPN++ and TransT with absolute gains of 4.6\% and 3.1\% in terms of AUC. The code and models will be released.
\end{abstract}

\section{Introduction}
\label{sec:introduction}
\IEEEPARstart{G}{iven} the initial position of an arbitrary object, visual object tracking is to  predict the position accurately and steadily in subsequent sequences. 
As a fundamental task in computer vision, tracking is widely used in many fields, such as video monitoring~\cite{tian2011video}, scene parsing~\cite{wang2023dionysus},
and robotic vision~\cite{sakagami2002intelligent,du2018unmanned}.
Substantial progress has been achieved, mainly owing to deep feature extracting~\cite{alexnet,inception,resnet,attention_is_all}, adaptive appearance modeling~\cite{kcf,atom,dimp,kys}, and correlation matching~\cite{siamesefc,siameserpn,ocean}.
Meanwhile, challenges remain, arising from polytropic appearance and motion state, distractor interference, background clutter, \etc.
Mainstream tracking approaches~\cite{siamesefc,siameserpn,dimp,ocean,transt} generate a Region of Interest (ROI) for each frame inference.
This region, also named search region, is usually an extended rectangle region centered on the previous predicted results or initial bounding box. 
Once acquired, the ROI will be sent to a deep neural network for object locating and scale estimating.
A prior condition of obtaining the ROI by this manner is the spatio-temporal continuity~\cite{bennett2004using} of physical objects.
Consequently, the previous location of the object can provide an indication of where the network should 
pay attention to~\cite{goturn}.
Capturing a target object within a search region rather than the whole image dramatically improves the efficiency.
Widespread, and long-term usage of this strategy, demonstrates its practicality and effectiveness in object tracking. 

\begin{figure}[!t]
	\begin{center}
		\includegraphics[width=0.48\textwidth]{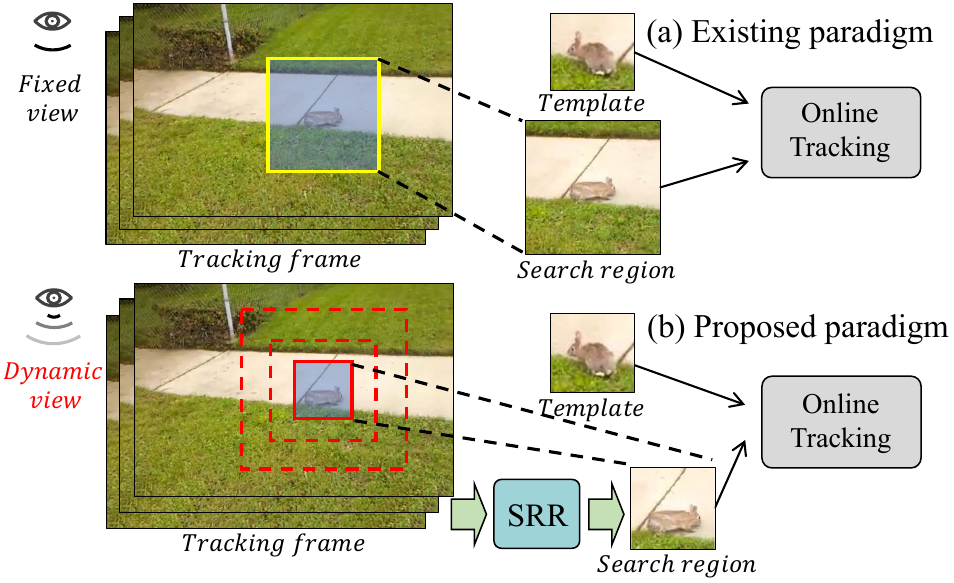}
		\caption{The paradigm of existing approaches (a) and the proposed (b). Existing tracking approaches adopt a fixed-size search region for target object capturing. In contrast, the proposed SRRT paradigm has a dynamic field of view, enabling the tracker a more robust and flexible tracking capability.
		}
		\vspace{-1mm}
		\label{fig:intro_diff}
	\end{center}
\end{figure}

Regardless of wide applications, the limitations of the above tracking paradigm remain.
In that paradigm, the search region of each frame is obtained by a fixed-size expansion on the previous predicted location.
Taking the object movement and tracker's estimating deviation into account, the expansion is usually several times the estimated target object area (\eg, 
$2^2$ times in GOTURN~\cite{goturn}, 
$4^2$ times in SiamRPN~\cite{siameserpn}, and $5^2$ times in STARK~\cite{stark}). 
With a fixed-size search region, trackers tend to lose the object because of the limited search region, conversely, be interfered with by distractors due to excessive range.
In particular, once fails, the tracker will experience difficulty recovering from drift due to the regional limitations of the fixed-size search region.
So, what size of search region is appropriate for stably and accurately capturing a target object?
This problem is worthy of exploration but ignored by researchers.

\begin{figure}[!t]
	\includegraphics[width=0.49\textwidth]{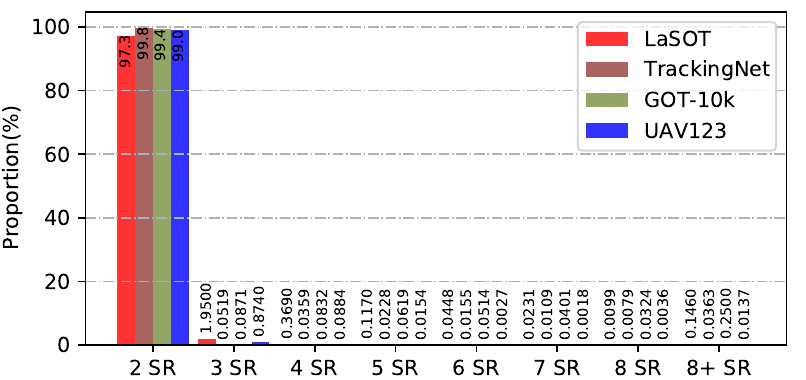}
	\caption{Minimum search region size distribution statistics of adjacent frames. 
		`$N$ $SR$': search region of $N^2$ times of previous object area.
		Across all above benchmarks, small search region ($2$$SR$) occupies a very high proportion, while extraordinarily  large search region exists at the same time.
	}
	\label{fig:intro_statis}
\end{figure}

With the above question, we first conduct probing experiments to investigate the necessity and benefit of optimizing the search region in object tracking.
Figure~\ref{fig:intro_statis} shows the minimum search region distribution of adjacent frames on four tracking benchmarks. 
We utilize the ground truth of the previous frame to calculate the minimum search region required in the current frame and obtain the minimum search region distribution of adjacent frames. 
As can be observed, surprisingly, 
the search region of $2^2$ times
occupies 97.3\%, 99.8\%, 99.0\%, and 99.4\% in LaSOT~\cite{lasot}, TrackingNet~\cite{trackingnet}, UAV123~\cite{uav}, and GOT-10k~\cite{got10k}, respectively.
This finding indicates that a tracker can sufficiently capture the target object in the next frame with a small-size search region at most times.
However, in some corner cases, an extraordinarily  large search region is required, such as $8^2$ times.
Although those cases are rare, neglecting them will result in a broken of trajectory chain. 
Furthermore, to verify the benefit of optimizing the search region,
we conduct an oracle experiment to gain the upper bound by feeding the optimal search region to the tracker. The optimal search region is gained by the ground truth for every frame and we take TransT~\cite{transt} as the base tracker. We are surprised to find that TransT improved from 64.9\% to 74.9\% in terms of AUC on LaSOT~\cite{lasot} benchmark. On the basis of the above probing analyses, we believe that the unsuitable search region greatly limits the performance of a tracker, and solving the tracking problem from the perspective of optimizing the search region can lead to potential enhancements. 

Inspired by the above observations, in this work, we propose a novel tracking paradigm, called Search Region Regulation Tracking (SRRT), which can dynamically generate search regions for tracking.
Compared to the conventional paradigm (Figure~\ref{fig:intro_diff} (a)),
SRRT (Figure~\ref{fig:intro_diff} (b)) applies a search region regulator (SRR) to allocate an optimal search region for the subsequent tracking components.
Taking advantage of the pre-perception of ROI, SRRT tracks a target object 
with a small size search region most of the time, and can flexibly switch to a larger search radius when encountering the cases such as object fast motion.
In addition, 
the variation in the appearance of objects during online tracking usually obscures the reference information from the initial frame.
We further propose a locking-state determined reference frame updating strategy, which can improve the modeling robustness for search region awareness without incurring additional computation overhead.
In addition, the proposed SRRT can be easily applied to existing trackers to improve tracking performance.
The main contributions of this work are summarized as follows:

\begin{itemize}
	\item We propose SRRT, a novel Search Region Regulation Tracking paradigm. Instead of adopting a fixed-size search region in all scenarios as with previous work practices, SRRT explores dynamically selecting an optimal search region for each frame, delivering higher performance while maintaining efficiency.
	\item We analyze the superiority of dynamically selecting a search region for visual object tracking and propose a simple yet effective pre-perception search region regulator module to estimate the optimal search region size.
	\item We propose a locking-state determined update strategy to improve the richness of the reference frame to adapt to the challenges caused by the variable object appearance.
	\item Extensive experiments on eight benchmarks demonstrate the effectiveness of the proposed method, especially on large-scale LaSOT, TrackingNet, and GOT-10k.
\end{itemize}

\section{Related Work}
\subsection{Visual Object Tracking}
Existing deep tracking algorithms can be roughly grouped into either online networks~\cite{eco,atom,dimp,prdimp,mdnet} or siamese networks~\cite{sint,siamesefc,siameserpn,siamrpnplusplus,siamfc++,siammask}. 
Online networks usually learn an online filter from the object examplar patches
to discriminate from the background.
These trackers sacrifice slightly in speed due to the existence of an online fine-tuning design.
Siamese trackers~\cite{siamesefc,siamfc++,siamrcnn,siamatten,zhou2021object} take the tracking problem as a template similarity matching problem.
To introduce more dynamics from the local and global search results, DeepMTA~\cite{wang2021dynamic}
proposes to build multi-trajectory tracking strategy, generating more candidates to represent and find the true target object. 
Although effective, modeling multiple tracking trajectories at the same time can lead to a significant loss of efficiency.
Then,  SiamTHN~\cite{siamthn} proposes to balance the weights of different channels to make the response maps more focused on the target region.
Recently, transformer-based trackers~\cite{transt,stark,trdimp,ostrack,mixformer,swintrack,tang2023learning} introduce transformer~\cite{attention_is_all} to the tracking framework, pushing the performance  to a new level. 
TransT~\cite{transt} proposes to replace the cross-correlation process with a Transformer fusion module, attention-based fusion is proven to be superior to local linear computing.
TransInMo~\cite{guo2022learning} enhances the target-perception representation for tracking by injecting the prior knowledge of the template patch into different stages of the feature extractor.	 
TransMDOT~\cite{chen2023cross} proposes a transformer-based multi-drone tracker that can model the association between multiple scenes from different shots. 
To fully mine temporal and spatial cues, Zhou $et~al.$~\cite{pstm} couple a proposed pixel-level spatio-temporal memory into the object tracking pipeline.
Recently, some works~\cite{ostrack,mixformer,swintrack} introduce pure transformer architecture for visual tracking, and the Siamese-based dual-stream network evolves into the one-stream paradigm.
The feature interaction between the template and the search region occurs in the backbone, 
thereby higher performance is brought by wealthier semantic interaction.
Some other works~\cite{vtuav,vipt} introduce auxiliary modal cues to enhance the target locating ability in complex scenarios.

However, these methods take tracking as a template classification or matching problem with a fixed local search region.
The appearance or matching model captures the target object in a fixed predefined ROI, thereby limiting the flexibility of the tracking algorithm, as described in Section~\ref{sec:introduction}.
In this work, we are committed to opening the limit of fixed search region, providing more efficient and flexible search region selection.

\subsection{Search Region Generation}
Generating a search region for the next frame input before the tracking process is a strategy that has been used by researchers for a long time.
This strategy is based on the prior information that the location and scale changement of the target object in the consecutive sequence is continuous. 
SiamFC~\cite{siamesefc} adds a margin to object the bounding box to generate a search region with a size of 255$\times$255 pixels, approximately $4^2$ times larger than the target object area. 
GOTURN~\cite{goturn} proposes to adopt a 0 mean Laplace distribution to model the smoothness of object motion and scale changes through space.
The search radius is twice the width and height of the predicted bounding box in the previous frame.
The tracker can better adapt to small movements, thereby achieving the continuity of tracking. 
To exploit scene information, KYS~\cite{kys} applies a $5^2$ times search region as large as the shape of the target object and combines the propagated dense localized vectors with the appearance model features to localize the target object.
SiamR-CNN~\cite{siamrcnn} is more robust to tracking drift by explicitly modeling the motion and interactions of all potential objects.
Despite obtaining excellent performance, the running speed of SiamR-CNN is extremely slow because it feeds the full-size image into the tracking network for perceiving all the candidate objects.
Other trackers also use a fixed search region,
\eg~$4^2$ times~\cite{siameserpn,siammask} and $5^2$ times~\cite{atom,dimp} the size of the previous object size.

\begin{figure*}[t]
	\centering
	\includegraphics*[width=0.9\textwidth]{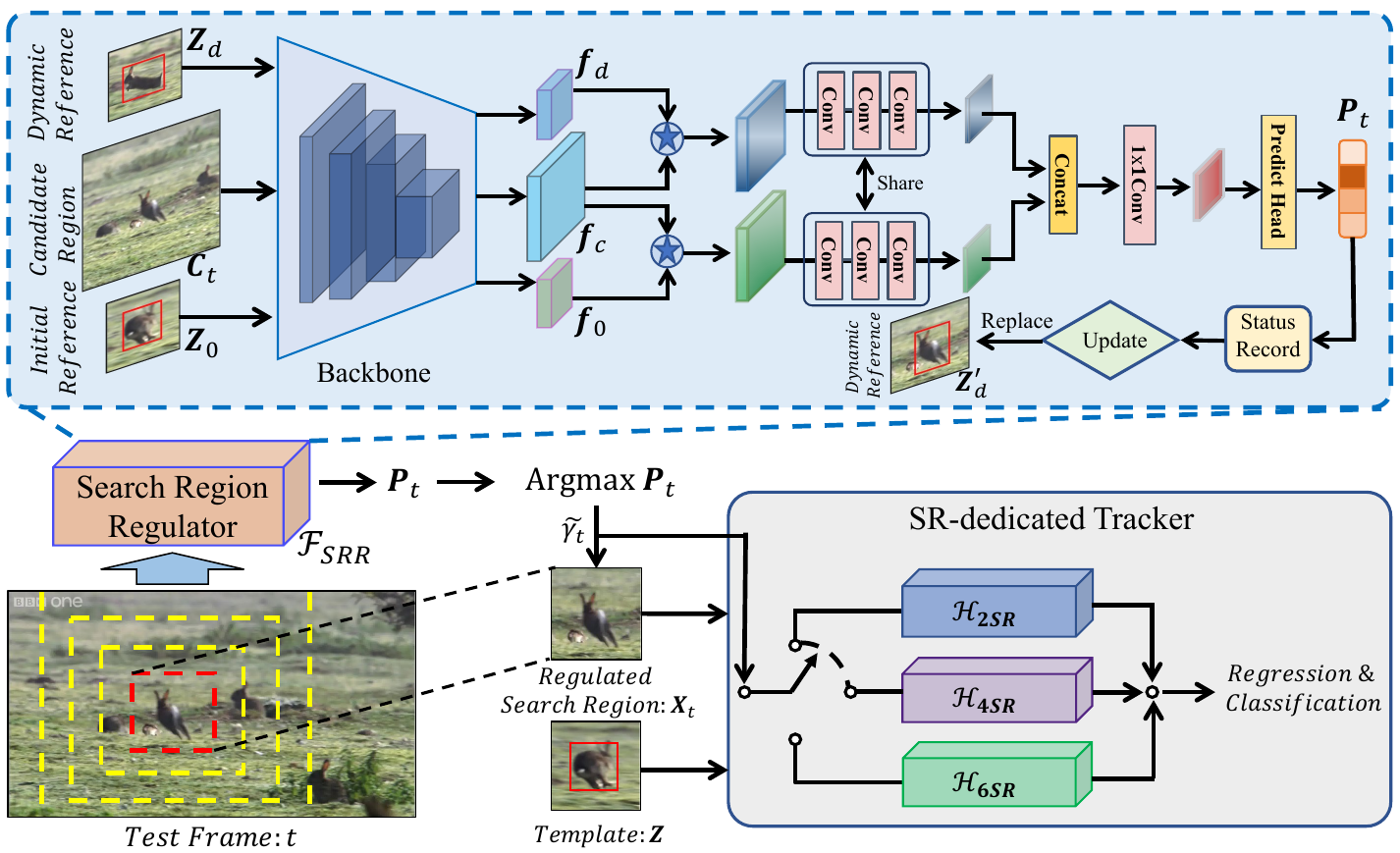}
	\caption{Overview of SRRT paradigm. Search Region Regulator (SRR) sweeps through the candidate region against initial and dynamic reference frames to generate a prediction of the desired search radius. Afterward, the search region of the current frame is cropped according to the predicted search radius, and the search region and template patches are fed to the corresponding SR-dedicated tracker to obtain the final tracking results.}
	\label{fig:method_main}
\end{figure*}

These methods employ different search regions, from $2^2$ times as large as the target object area to the whole image.
However, no matter how large a search region is employed, limitations remain.
A small search region brings advantages in speed, while may easily lose the target object, \eg,~when the target object moves quickly.
A large search region increases the search scope, but it also needs to face the increased distractors. Therefore, a complex identity discrimination or association module needs to be designed to filter the interference.
To deal with occlusion in long-term tracking, the tracking-learning-detection (TLD) method~\cite{tld} combines a local tracker and a global re-detector. 
{
	FreeAnchor~\cite{freeanchor} shifts hand-crafted anchor assignment to a learnable anchor matching, and formulates the training of the detection model as a maximum likelihood estimation procedure.
	MAT~\cite{mat} introduces the multi-anchor mechanism to predict multiple-anchor regions and gains ensemble results, the sizes of search regions (anchors) are fixed and identical.
	On the contrary, our work implements a more flexible online tracking paradigm, which predicts an optimal search region for each frame, improving performance while being concise.
}


\section{Search Region Regulation Tracking}

In this section, we describe how our method (Figure~\ref{fig:method_main}) achieves online tracking with a regulated search region. 
We initially revisit the standard search region generation approach and analyze its limitations (Section~\ref{sec:3.1}). 
Subsequently, we introduce the formulation of our search region regulation (Section~\ref{sec:3.2}) and describe the components and training process of our model in detail (Section~\ref{sec:3.3}).
Finally, the pipeline of the SRRT framework is described in Section~\ref{sec:3.4}.

\subsection{Conventional Search Region Generation} 
\label{sec:3.1}
Existing trackers~\cite{goturn,siameserpn,atom} generate an ROI for the appearance modeling of each frame.
This manner is based on the accuracy of the previous prediction. Because it is simple and effective, this manner has dominated visual object tracking until now. 
For one particular target object, most trackers exploit its location ${\boldsymbol L_{t-1}}: (c_x^{t-1},c_y^{t-1})$ 
and scale ${\boldsymbol S_{t-1}}:(h^{t-1},w^{t-1})$ in the $(t-1)$th frame and directly yield the search region ${\boldsymbol X_{t}} \in \mathbb{R}^{H_x \times W_x \times 3}$ for current frame tracking, 
\begin{equation}
	{\boldsymbol X_{t}}=crop(\boldsymbol L_{t-1}, \gamma \boldsymbol S_{t-1} ),
	\label{con_SRR}
\end{equation}
where $crop(\cdot)$ represents cropping the region centered on ${\boldsymbol L_{t-1}}$ to generate the corresponding search region.
The search radius factor $\gamma$ is set to a fixed value (\eg,~$\gamma=4$). 
Trackers perform appearance modeling and matching in this region that is $\gamma^{2}$ times the $(t-1)$th frame target object area.

In this work, we argue that this hidebound manner limits the tracker's performance.
Its limitations can be mainly concluded as follows: 
(\textbf{\romannumeral1}) Applying a fixed search region for every video, every frame, can barely be aware of the existence of the target object, which means trackers may lose the target object caused by some factors, \eg,~fast motion, or be interfered by distractors due to excessive search region range.  
(\textbf{\romannumeral2}) Once tracking fails, the tracker lacks adaptability, suffering from difficulty to rediscover the target object due to limited search region.
(\textbf{\romannumeral3}) To alleviate the phenomenon of target object loss (\eg, $5^2$ times), most trackers adopt a large search region, which reduces the efficiency of online tracking.

\subsection{Learning Search Region Regulation}
\label{sec:3.2}
We design a pre-perception search region regulator to estimate a proper search region dynamically.
The essential process in SRR is to estimate the search radius factor $\widetilde{\gamma}_t \in \{\widetilde{\gamma}_j\}_{j=1}^{M}$ for each frame through given information according to, 
\begin{equation}
	\widetilde{\gamma}_t={\rm Argmax}(\mathscr{F}_{SRR}(\boldsymbol Z_0, \boldsymbol C_t)),
	\label{gamma_eq}
\end{equation}
where $\mathscr{F}_{SRR}(\cdot)$ donates the learned generic function for search region regulation.
$\boldsymbol Z_0 \in \mathbb{R}^{H_0 \times W_0 \times 3}$ represents target reference patches, and $\boldsymbol C_t \in \mathbb{R}^{H_c \times W_c \times 3}$  represents the candidate region patches of the current frame for the search region regulation function. Specifically, candidate region $\boldsymbol C_t$ is similar to the search region but only used by SRR.
The learned function $\mathscr{F}_{SRR}(\cdot)$  computes the probability for search radius factor $\widetilde{\gamma}_j$ by using the reference target $\boldsymbol Z_0$ and the candidate region $\boldsymbol C_t$ patches.
In addition, ${\rm Argmax}(\cdot)$ denotes selecting the optimal search radius factor $\widetilde{\gamma}_t$.
Figuratively, $\mathscr{F}_{SRR}(\cdot)$ attempts to simulate the human behavior
when tracking a target object, that we first glance to determine its approximate location before finding its extract position.
Therefore, the tracker can obtain an optimal search region at the frame level, improving the adaptability for variable target movement states.
In this work, $\mathscr{F}_{SRR}(\cdot)$ is implemented as a neural network, which can learn general mapping by a large amount of data.
Moreover, search region regulation is independent of the subsequent tracking process.
Thus, it is a plug-and-play module, that can be flexibly integrated into existing trackers. 

\subsection{Search Region Regulator}
\label{sec:3.3}
{\noindent \textbf{Model Design.}}
We present the structure of our search region regulation (SRR) module in Figure~\ref{fig:method_main}. The pipeline is described in an online tracking process.
SRR employs a Siamese-based architecture to match the reference target object from the candidate region.
We adopt dual reference patches, an initial reference $\boldsymbol Z_0$ 
sampling from the ground-truth bounding box in the initial frame,
and a dynamic reference $\boldsymbol Z_d \in \mathbb{R}^{H_d \times W_d \times 3}$ updating by online tracking results. These two patches are used for robust search region regulation on a candidate region $\boldsymbol C_t$, which is $6^2$ times of the previous object area. 
A ResNet-50~\cite{resnet} backbone is employed for feature extraction from reference and candidate region patches. Reference features ($\boldsymbol f_0$, $\boldsymbol f_d$) and candidate features $\boldsymbol f_c$ are correlated by the depth-wise correlation operation~\cite{siammask,siamrpnplusplus}. 
The correlated features are first sent to a parameter-sharing convolution block. The produced two feature maps are concatenated and then sent to a $1 \times 1$ convolutional layer used as weighted summation.
Lastly, a prediction head consisting of a three-layer MLP outputs the probability $\boldsymbol{P}_t$ for different search radius factors.
The search region regulation with a dynamic reference can be described as follows:
\begin{equation}
	\widetilde{\gamma}^*_t={\rm Argmax}(\mathscr{F}_{SRR}(\boldsymbol Z_0, \boldsymbol Z_d, \boldsymbol C_t)),
	\label{SRR_eq}
\end{equation}
Compared with the conventional Siamese trackers, the feature fusion of SRR is simple yet efficient,
because a tracker needs to complete accurate target object locating, whereas SRR only needs to predict the category of the search region.

\begin{figure}[t]
	\centering
	\includegraphics[width=0.49\textwidth]{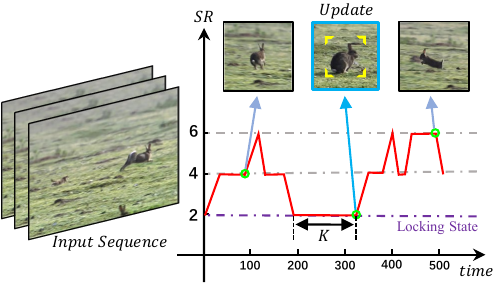}
	\caption{Intuitive presentation of locking-state determined update strategy. During online tracking at the proposed SRRT paradigm, if the SRR assigns a minimum search region (\eg, $2^2$ times in this work) for the tracker in consecutive $K$ frames, we term it target-locking and the current frame will be updated as the dynamic reference frame. This strategy allows the model to gain enhanced adaptability to the changes in the target appearance without introducing additional update network modules.
	}
	\label{fig:LDU_vis}
\end{figure}

{\noindent \textbf{Locking-state Determined Update.}}
Dynamic reference frame is beneficial to improve the adaptation to the target's appearance changement during online tracking.
Some existing approaches~\cite{fear,stark} also use a dual-template representation for target model adaptation. 
Different from these approaches, we propose a locking-state determined update strategy that does not require any extra network or computing to be used as update judgments.
On the contrary, as shown in Figure~\ref{fig:method_main}, we record the search region selected for each frame in a sequence. 
When the SRRT selects the smallest search region for consecutive frames, we name this situation as the target “locking-stage", which can be expressed by,
\begin{equation}
	\widetilde{\gamma}^*_t = \widetilde{\gamma}^*_{t-1} = \ldots = \widetilde{\gamma}^*_{t-K+1} = r_{min},
\end{equation}
where $K$ is the locking-state threshold and $r_{min}$ is the minimum search radius ($r_{min}=2$ in this work). 
Figure~\ref{fig:LDU_vis} showcases an intuitive presentation of the proposed locking-state determined update strategy.
Once this situation appears, the model updates the dynamic reference $\boldsymbol Z_d$ with 
$\boldsymbol X_t$ 
from the current frame. 
Another thing to note is that our locking-state determined update strategy is designed for the SRR, not the base tracker, so our SRR module is completely independent of the SR-dedicate base trackers.

{\noindent \textbf{Training Approach.}}
For training our search region regulator, we randomly sample three frames in the same video sequence with an interval of no more than 100 frames to generate the reference and candidate region patches.
The initial reference frame and candidate region are always from the beginning or end of a sequence. Two reference patches and one candidate region form a sampling pair.
The candidate patches are sampled  with different search region distributions for model training.
The location $\boldsymbol {L}: \{c_x, c_y\}$ and scale $\boldsymbol {S}: \{h, w\}$ of the sampling candidate region are determined by the equations:
\begin{gather}
	\{h, w\}=\{h_{gt}, w_{gt}\} \times \gamma_{train} \times  e^{\delta_S}, \\
	\{c_x, c_y\}=\{c_{x}^{gt}, c_{y}^{gt}\} + \frac{h+w}{2} \times \delta_C,
\end{gather}
where $h_{gt}$, $w_{gt}$, $c_{x}^{gt}$, $c_{y}^{gt}$ are obtained from the ground-truth bounding box, and we set the search radius $\gamma_{train}=6$.
$\delta_S$ and $\delta_C$ are factors for scale and location jitters as data augmentation. 
Specially, we apply variables  $\delta_S$ and $\delta_C$ to form the sample pairs with different categories of search regions. The training set $T=\{(\boldsymbol {C}_i, \boldsymbol{Z}_0^i, \boldsymbol{Z}_d^i, Y_i)\}_{i=1}^{n}$ consists of three input patches and search region category annotation $Y_i$.
In our experiments, the sampling ratio of search region categories is set to $2\ SR: 4\ SR: 6\ SR : 8\ SR = 1 : 1 : 1 : 1$, where $8\ SR$ samples are used to simulate the situation that the target object disappears in the candidate region.
For training the search region regulator network, we merely use a cross-entropy loss $L_c$, as follows:
\begin{equation}
	L_c=-\frac{1}{N}\sum\nolimits_{i=1}^{N} \sum\nolimits_{j=1}^{M} y_{i,j} \log \left(p_{i,j}\right)
\end{equation}
where $M$ denotes the number of search region categories, and $N$ is the size of a mini-batch. $y_{i,j}$ and $p_{i,j}$ represent the ground truth and predicted probability for the $j$th category search region of the $i$th sample, respectively.
Note that the training process of SRR is independent of the SR-dedicated trackers, thus it does not need to match a specific tracker.	

\subsection{SRRT Pipeline}
\label{sec:3.4}
In this subsection, we describe the pipeline of the SRRT framework for online tracking.
As shown in Figure~\ref{fig:method_main}, reference patches $(\boldsymbol{Z}_0,\boldsymbol{Z}_d)$ and candidate region patches $\boldsymbol{C}_t$ are sent to the search region regulation module, and an optimal search radii factor $\widetilde{\gamma}^*_t$ is predicted by the search region regulator (SRR). 
Then, the search region patch is cropped in terms of $\widetilde{\gamma}^*_t$.
In order to handle search regions with different search radius, SR-dedicated trackers are employed. 
We take three SR-dedicated trackers $\mathscr{H}_{2SR}$, $\mathscr{H}_{4SR}$, and $\mathscr{H}_{6SR}$, 
which are responsible for processing the input search region of $2^2$, $4^2$, and $6^2$ times of the target object area, respectively.
Especially, $\mathscr{H}_{6SR}$ is employed for handling search regions equal to or larger than $6^2$ times the target area. 
The corresponding SR-dedicated tracker is selected in terms of $\widetilde{\gamma}^*_t$ for the current frame. 
Finally, the regression and classification results are output by corresponding prediction heads.
SRRT paradigm flexibly provides suitable search regions for different tracking scenarios and combines the power of different SR-dedicated trackers through online dynamic switching.

\section{Experiments}

\begin{table*}[!t]
	\centering
	\renewcommand\arraystretch{1.2}
	\normalsize 
	\caption{\ \ \ \ \ State-of-the-art comparison on LaSOT~\cite{lasot}, TrackingNet~\cite{trackingnet}, GOT-10k~\cite{got10k}, UAV123~\cite{uav}, and LaSOT$_{ext}$~\cite{lasotext} test benchmarks. \ \ \ \ \  
		$*$ denotes the results provided by us, and the SRRSiamRPN++ is based on SiamRPN++$^{*}$. SRRT shows consistent improvements on all test benchmarks. }
	\begin{tabular}{
			c@{\hspace{3pt}}|c@{\hspace{3pt}}c@{\hspace{3pt}}c@{\hspace{3pt}}c@{\hspace{3pt}}c@{\hspace{3pt}}c@{\hspace{3pt}}c@{\hspace{3pt}}c@{\hspace{3pt}}c@{\hspace{3pt}}c@{\hspace{3pt}}c@{\hspace{3pt}}c@{\hspace{3pt}}c@{\hspace{3pt}}c}
		\toprule
		\multirow{2}{*}{Methods} &   \multicolumn{3}{c}{LaSOT} & \multicolumn{3}{c}{TrackingNet} & \multicolumn{3}{c}{GOT-10k} & \multicolumn{2}{c}{UAV123} & \multicolumn{3}{c}{LaSOT$_{ext}$}\\
		\cline{2-15}
		
		& AUC & P$_{Norm}$ & P & AUC & P$_{Norm}$ & P
		& AO & SR$_{50}$ & SR$_{75}$ & AUC & P & AUC & P$_{Norm}$ & P\\
		\midrule      
		SiamFC
		&33.6&42.0&33.9&57.1&66.3 &53.3 &34.8 &35.3 &9.8 &48.5 &69.3&23.0&31.1&26.9  \\
		MDNet
		&39.7 &46.0 &37.3 &60.6 &70.5 &56.5 &29.9 &30.3 &9.9 &52.8 &-&27.9&34.9&31.8 \\
		ECO
		&32.4 &33.8 &30.1 &55.4 &61.8 &49.2 &31.6 &30.9 &11.1 &52.5 &74.1&22.0&25.2&24.0 \\
		SiamRPN++
		&49.6 &56.9 &49.1 &73.3 &80.0 &69.4 &51.7 &61.6 &32.5 &61.0 &80.3&34.0&41.6&39.6\\
		ATOM
		&51.5 &57.6 &50.5 &70.3 &77.1 &64.8 &55.6 &63.4 &40.2 &64.3 &-&37.6&45.9&43.0 \\
		DiMP
		&56.9 &65.0 &56.7 &74.0 &80.1 &68.7 &61.1 &71.7 &49.2 &65.4 &-&39.2&47.6&45.1 \\
		SiamFC++
		&54.4 &62.3 &54.7 &75.4 &80.0 &70.5 &59.5 &69.5 &47.9 &- &-&-&-&- \\
		OCEAN
		&56.0 &65.1 &56.6 &- &- &- &61.1 &72.1 &47.3 &- &-&-&-&- \\
		PrDiMP
		&59.8 &68.8 &60.8 &75.8 &81.6 &70.4 &63.4 &73.8 &54.3 &68.0 &-&-&-&- \\
		SiamR-CNN
		&64.8&72.2&-&81.2 &85.4 &80.0 &64.9 &72.8 &59.7 &64.9&{{83.4}}&-&-&- \\ 
		STMTracker
		&60.6 &69.3 &63.3 &80.3 &85.1 &76.7 &64.2 &73.7 &57.5 &64.7 &-&-&-&- \\
		TrDiMP
		&63.9 &- &61.4 &78.4 &83.3 &73.1 &67.1 &77.7 &58.3 &67.5 &-&-&-&- \\
		STARK-ST50
		&66.4&-&71.2&81.3&86.1&-&{68.0}&{77.7}&{62.3}&69.1&-&-&-&- \\
		KeepTrack
		&67.1 &{{77.2}} &70.2 &-&-&-&-&-&-&{{69.7}} &-&{48.2}&-&- \\
		SparseTT
		&66.0 &74.8 &70.1  &81.7 &86.6 &79.5 &69.3 &{79.1} &{63.8} &70.4 &-&-&-&- \\
		CSWinTT
		&66.2 &75.2 &70.9  &81.9 &{86.7} &79.5 &{69.4} &78.9 &{65.4} &{70.5} &{90.3} &-&-&- \\
		SBT-base
		&65.9 &- &70.0 &-&-&-&{69.9}&{80.4}&{63.6}&- &-&-&-&- \\
		TrDiMP\_PAF &64.4&-&67.1&78.1&-&74.3&69.6&80.7&-&-&-&-&-&-\\
		SiamTactic\_P &65.9&-&73.8&-&-&-&-&-&-&69.1&90.9&-&-&-\\
		GTELT
		&{67.7} &- &{73.2} &{82.5}&{86.7}&{81.6}&-&-&-&- &-&45.0&{54.2}&{52.4} \\
		\midrule 
		SiamRPN++$^{*}$
		&{52.3} &{59.1} &{52.4} &{75.1} &{80.5} &{71.0} &{57.3} &{67.1} &{47.0} &{62.1} &{78.5}&35.4&40.0&38.1\\
		TransT
		&{64.9} &{73.8} &{69.0} &{81.4} &{86.7} &80.3 &67.1 &76.8 &60.9 &{69.1} &- &45.1&51.3&51.2\\
		\midrule 
		\textbf{SRRSiamRPN++}  
		&56.9\textbf{\dt{+4.6}} &64.0 &57.1 &76.0\textbf{\dt{+0.9}} &81.3 &71.9 &58.3{\textbf{\dt{+1.0}}} &68.0 &46.8 &62.8\textbf{\dt{+0.7}} &78.8 &37.0\textbf{\dt{+1.6}} &41.7&40.0\\
		\textbf{SRRTransT}~ 
		&68.0\textbf{\dt{+3.1}} &{76.9} &{{72.4}} &82.1\textbf{\dt{+0.7}} &{{87.2}} &{{80.4}} &67.7\textbf{\dt{+0.6}} &77.1 &61.5 &71.1\textbf{\dt{+2.0}} &88.5&47.4\textbf{\dt{+2.3}}&{54.0}&{54.1} \\
		\bottomrule
	\end{tabular}
	\label{tab:compare_sota}
\end{table*}

\subsection{Implementation Details}
SRRT is implemented with the Pytorch~\cite{pytorch} library. 
We verify the effectiveness of SRRT on eight benchmarks, including LaSOT~\cite{lasot}, TrackingNet~\cite{trackingnet}, GOT-10k~\cite{got10k}, UAV~\cite{uav}, LaSOT$_{ext}$~\cite{lasotext}, NFS~\cite{nfs}, OTB100~\cite{otb2015}, and VOT-LT2020~\cite{vot20}.
We take two classic trackers, SiamRPN++~\cite{siamrpnplusplus} and TransT~\cite{transt} as our base tracker. SiamRPN++~\cite{siamrpnplusplus} is a representative work of the Siamese-based methods, and TransT~\cite{transt} is a recent transformer-based method. 
We choose them to demonstrate the effectiveness of the proposed SRRT paradigm on both the CNN and Transformer models.

{\noindent \textbf{Offline Training.}}
The training set of SRRT includes the training splits of LaSOT~\cite{lasot}, GOT-10k~\cite{got10k}, COCO~\cite{coco}, and TrackingNet~\cite{trackingnet}. 
Following the convention~\cite{stark}, we remove 1k forbidden sequences from the training set of GOT-10k.
The reference patches are resized to 128$\times$128 pixels, and the candidate patches are resized to 384$\times$384 pixels.
The SRR module employs a ResNet-50~\cite{resnet} backbone initialized with ImageNet~\cite{imagenet} pre-trained parameters. 
AdamW~\cite{adamw} optimizer is adopted during model training.
The learning rate is set to 1e-3 and decayed by 10$\times$ at every 30 epochs; 
The weight decay is set to 1e-4. 
The SRR model is trained on two Nvidia RTX 2080Ti GPUs for 90 epochs with a batch size of 32. Each epoch includes 50,000 sample pairs.

For the SR-dedicated base tracker, we trained two other TransT~\cite{transt} models that adopted a search region of $2^2$ times and $6^2$ times, respectively. We use the original TransT parameters as our $4^2$ times search region model.
The template patches are resized to 128$\times$128 pixels, and the search regions of $2^2$, $4^2$, and $6^2$ times are resized to 128$\times$128, 256$\times$256, and 384$\times$384 pixels, respectively.
Other training settings for different search regions are consistent with the original version.
For the convenience of experiments, SimaRPN++\cite{siamrpnplusplus} was reproduced with different search regions.
The model architecture is maintained the same as that in the original paper except for training datasets and the optimizer. 
We use the same training datasets and optimizer as the TransT.
The SR-dedicated models are trained on four Nvidia RTX Titan GPUs.

{\noindent \textbf{Online Tracking. }}
During inference, only window penalty post-processing is applied; it is also used in SiamRPN++\cite{siamrpnplusplus} and TransT~\cite{transt} to reweigh the classification score maps. In addition, our SRRT framework does not have any online fine-tuning modules. 

\begin{figure*}[t]
	\centering
	\includegraphics[width=1.0\textwidth]{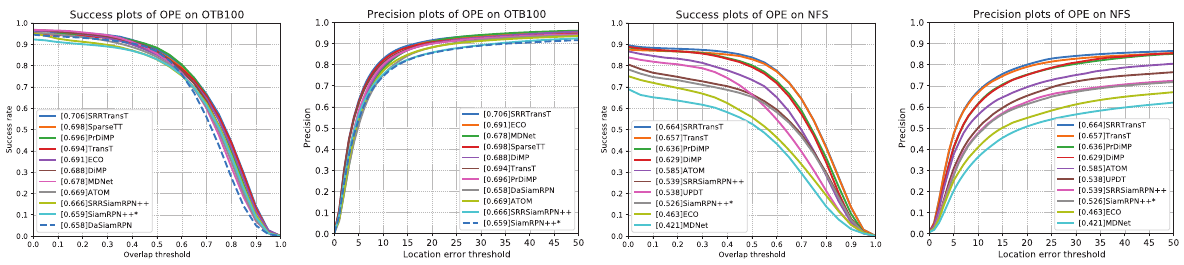}
	\caption{State-of-the-art comparison on OTB100 and NFS. SRRT consistently improves the performance of baselines.
		Best viewed in color with zoom-in. 
	}
	\label{fig:otb_nfs}
\end{figure*}

\subsection{State-of-the-art Comparison}
We compare SRRT on seven widely used benchmarks with state-of-the-art trackers including SiamFC~\cite{siamesefc}, MDNet~\cite{mdnet}, ECO~\cite{eco}, SiamRPN++~\cite{siamrpnplusplus}, ATOM~\cite{atom}, DiMP~\cite{dimp}, SiamFC++~\cite{siamfc++},
OCEAN~\cite{ocean}, PrDiMP~\cite{prdimp}, SiamR-CNN~\cite{siamrcnn}, STMTracker~\cite{stmtrack}, TrDiMP~\cite{trdimp}, TransT~\cite{transt},  STARK~\cite{stark}, KeepTrack~\cite{keeptrack},  SparseT~\cite{sparsett}, CSWinTT~\cite{cswintt}, SBT~\cite{SBT}, TrDiMP\_PAF~\cite{li2023part}, SiamTactic\_P~\cite{xu2023toward}, and GTELT~\cite{zhou2022global}. The test benchmarks include LaSOT~\cite{lasot}, TrackingNet~\cite{trackingnet}, GOT-10k~\cite{got10k}, UAV123~\cite{uav}, LaSOT$_{ext}$~\cite{lasotext}, OTB100~\cite{otb2015}, and NFS~\cite{nfs}.
In addition, a long-term tracking benchmark VOT-LT2020~\cite{vot20} is employed to verify the stability and superiority of SRRT on more challenging long-term tracking sequences.

{\noindent \textbf{LaSOT. }}
LaSOT~\cite{lasot} is a large-scale single-object tracking benchmark,
which consists of 1400 sequences.
It contains many long sequences and challenging scenes of distractors. 
The results in terms of success (AUC), normalized precision (P$_{Norm}$), and precision (P) are shown in Tab.~\ref{tab:compare_sota}.
Our SRRSiamRPN++ outperforms SiamRPN++ with a large margin, 4.6\% and 4.9\% on AUC and P$_{Norm}$, respectively.
These improvements clearly show the benefits of regulating an adaptive search region for real-world scene tracking.
For transformer-based tracker TransT, SRRTransT still boosts the baseline's performance to 68.0 in terms of AUC, which is an improvement of 3.1\%. 
SRRTransT outperforms the current state-of-the-art trackers, \eg,~GTELT~\cite{zhou2022global} and SiamTactic\_P~\cite{xu2023toward},
even though the base model TransT has a gap between them.

{\noindent \textbf{TrackingNet. }}
TrackingNet~\cite{trackingnet} is a large-scale tracking dataset with a test split of 511 sequences containing rich distributions of object classes. 
The performance evaluation is provided on an online official server following the one-pass evaluation protocol. 
Tab.~\ref{tab:compare_sota} shows the results.
Our SRRTransT achieves the best performance, namely, 82.1\%, 87.2\%, and 80.4\% for AUC, P$_{Norm}$, and P, respectively. Our SRRT paradigm boosts TransT and SimaRPN++ consistently, increasing AUC from 81.4\% and 75.1\% to 82.1\% and 76.0\%.

\begin{table*}[!h]
	\centering
	\fontsize{8pt}{3.5mm}\selectfont
	\caption{State-of-the-art comparison on VOT-LT2020~\protect\cite{vot20}. The methods for comparison are from VOT2020 challenge~\cite{vot20}.} 
	\setlength{\tabcolsep}{1.5mm}{
		\begin{tabular}{c|cccccc|cc|cc}
			\toprule
			&SPLT &ltMDNet &SiamDW\_LT &RLT\_DiMP &CLGS &LTMU\_B &SiamRPN++* &TransT &\textbf{SRRSiamRPN++} &\textbf{SRRTransT} \cr
			\midrule
			F-score  &52.6 &57.4&65.6&67.0 &67.4 &69.1&56.5&66.0& 57.0  (\bf{+0.5}) & 68.0 (\bf{+2.0})  \\
			Pr  &63.7 &64.9&67.8&65.7  &73.9 &70.1&58.7&69.9& 65.1& 68.7  \\
			Re  &44.8 &51.4&63.5&68.4  &61.9 &68.1&54.4& 62.6 &50.7 & 67.2 \\
			\bottomrule
		\end{tabular}
	}
	\label{tab:vot2020lt}
\end{table*}

{\noindent \textbf{GOT-10k. }}
GOT-10k~\cite{got10k} is a tracking dataset that broadly covers 560 classes of common outdoor moving objects, the training and testing sets are not intersected.
We submit results to its online evaluation server. Average overlap (AO) is employed for performance measure. As shown in Tab.~\ref{tab:compare_sota}, 
SRRTransT achives an AO score of 67.7\%, which is 
a better performance than the baseline. For SRRSiamRPN++, the AO score increases to 58.3\%. The results show that our SRRT can generalize effectively on different scenarios and challenges.

{\noindent \textbf{UAV123. }}
UAV123~\cite{uav} contains 123 low-altitude aerial videos with 9 object categories, the total frames are more than 112K.
Similar to LaSOT, it contains long sequences and distractor scenes, which can be challenging for robust object tracking. 
As shown in Tab.~\ref{tab:compare_sota},
SRRT boosts the performance of the baselines. 
It is worth noting that SRRTransT achieves the state-of-the-art performance with an AUC of 71.1\%.

{\noindent \textbf{LaSOT$_{ext}$.}}
LaSOT$_{ext}$~\cite{lasotext} is an extension of the LaSOT benchmark, containing 150 videos of 15 object categories, which is more challenging than LaSOT. As shown in Tab.~\ref{tab:compare_sota}, SRRT paradigm also demonstrates its superiority on LaSOT$_{ext}$ benchmark. SRRTransT and SRRSiamRPN++ outperform their baselines by 2.3\% and 1.6\% in terms of AUC, respectively. Our SRRTransT achieves  a competitive AUC (47.4\%) with the current best method KeepTracker~\cite{keeptrack} while having a 2$\times$ faster (41.8 $vs$ 18.3 $fps$) speed.

{\noindent \textbf{OTB100 and NFS.}}
OTB100~\cite{otb2015} and NFS~\cite{nfs} are two commonly used small-scale benchmarks both including 100 videos. We report the success plots over all videos, and the results are shown in Figure~\ref{fig:otb_nfs}. On OTB100, our SRRTransT obtains the best AUC score of 70.6\%, outperforming the previous best method PrDiMP by 1\%. SRRTransT and SRRSiamRPN boost their baselines by a relative gain of 1.7\% and 1.1\%. On NFS, SRRTransT and SRRSiamRPN achieve 66.4\% and 53.9\% of AUC score, which outperform the baselines relatively by 1.1\% and 2.5\%.
These results verify that SRRT also generalizes well in small-scale benchmarks.

{\noindent \textbf{VOT-LT2020.}}
VOT-LT2020~\cite{vot20} contains 50 long-term video sequences. Compared to other short-term benchmarks, VOT-LT2020 includes more challenging scenarios in which objects disappear and reappear frequently. 
VOT-LT2020 employs F-score=$\frac{2 P r R e}{P r+R e}$ to rank the performance of the trackers, where $Pr$ and $Re$ denote precision and recall, respectively.
As shown in Table~\ref{tab:vot2020lt}, SRRSiamRPN++ delivers a large performance improvement to the baseline method, 
increasing 4.4\% on F-score.
Besides, SRRTransT achieves an F-score of 68.0\%, which is higher than its base tracker. It is worth noting that the improvement is more remarkable in the recall score (from 62.6\% to 65.6\% on SRRTransT, from 44.8\% to 50.7\% on SRRSiamRPN++),  which indicates the favorable  robustness and re-detection capabilities of our SRRT.

\subsection{Ablation Studies}
In this subsection, we provide an ablation study analyzing our SRRT from various aspects with TransT~\cite{transt} as the base SR-dedicated tracker, and large-scale benchmark LaSOT~\cite{lasot} is used for evaluation.

{\noindent \textbf{Component-wise Analysis.}}
The component-wise study results are shown in Tab.~\ref{tab:com}.
\#NUM \textcircled{1} is the baseline and
\#NUM \textcircled{3} is our SRRT.
In \#NUM \textcircled{2}, after applying the search region regulation, the tracker achieves 2.2\%, 2.1\%, and 2.4\% 
improvement in AUC, P$_{Norm}$, and P, respectively. The improvement proves the effectiveness of the search region regulation paradigm because it can select a suitable search region flexibility during online tracking. 
Moreover, we apply the locking-state determined update (LDU) strategy for dynamic reference frame update. As shown in \#NUM \textcircled{3},
with dynamic reference updating, the search region regulator gains the ability to adapt to variations in target appearance,
thus can estimate the optimal search radius more accurately, further improving the final tracking performance. The AUC, P$_{Norm}$, and P are boosted to 68.0\%, 76.9\%, and 72.4\%, respectively, achieving a 
level of state-of-the-art performance.

\begin{table}[h]
	\centering
	\caption{Component-wise analysis of the proposed model. AUC, P$_{Norm}$, and P  results demonstrate the importance of each component in our framework. }
	\begin{tabular}{@{\hspace{-1pt}}c@{\hspace{-1pt}} | @{\hspace{-1pt}}c@{\hspace{-1pt}} @{\hspace{-1pt}}c@{\hspace{-1pt}}  @{\hspace{-1pt}}c@{\hspace{-1pt}} | @{\hspace{-1pt}}c@{\hspace{-1pt}} @{\hspace{-1pt}}c@{\hspace{-1pt}}  @{\hspace{-1pt}}c@{\hspace{-1pt}}}
		\toprule
		~~\#NUM~~ &~~Base~~ &~~SRR~~ &~~LDU~~ & ~~AUC~~ & ~~P$_{Norm}$~~  & ~~P~~ \\
		\midrule
		\textcircled{1} & \checkmark & & & 64.9 & 73.8  & 69.0  \\
		\textcircled{2}  & \checkmark &~~\checkmark  & &67.1  & 75.9 & 71.4 \\
		\textcircled{3} & \checkmark &~~\checkmark &\checkmark & 68.0  & 76.9 & 72.4 \\
		\bottomrule
	\end{tabular}
	\label{tab:com}
\end{table}

\begin{figure*}[t]
	\centering
	\includegraphics*[width=1.0\textwidth]{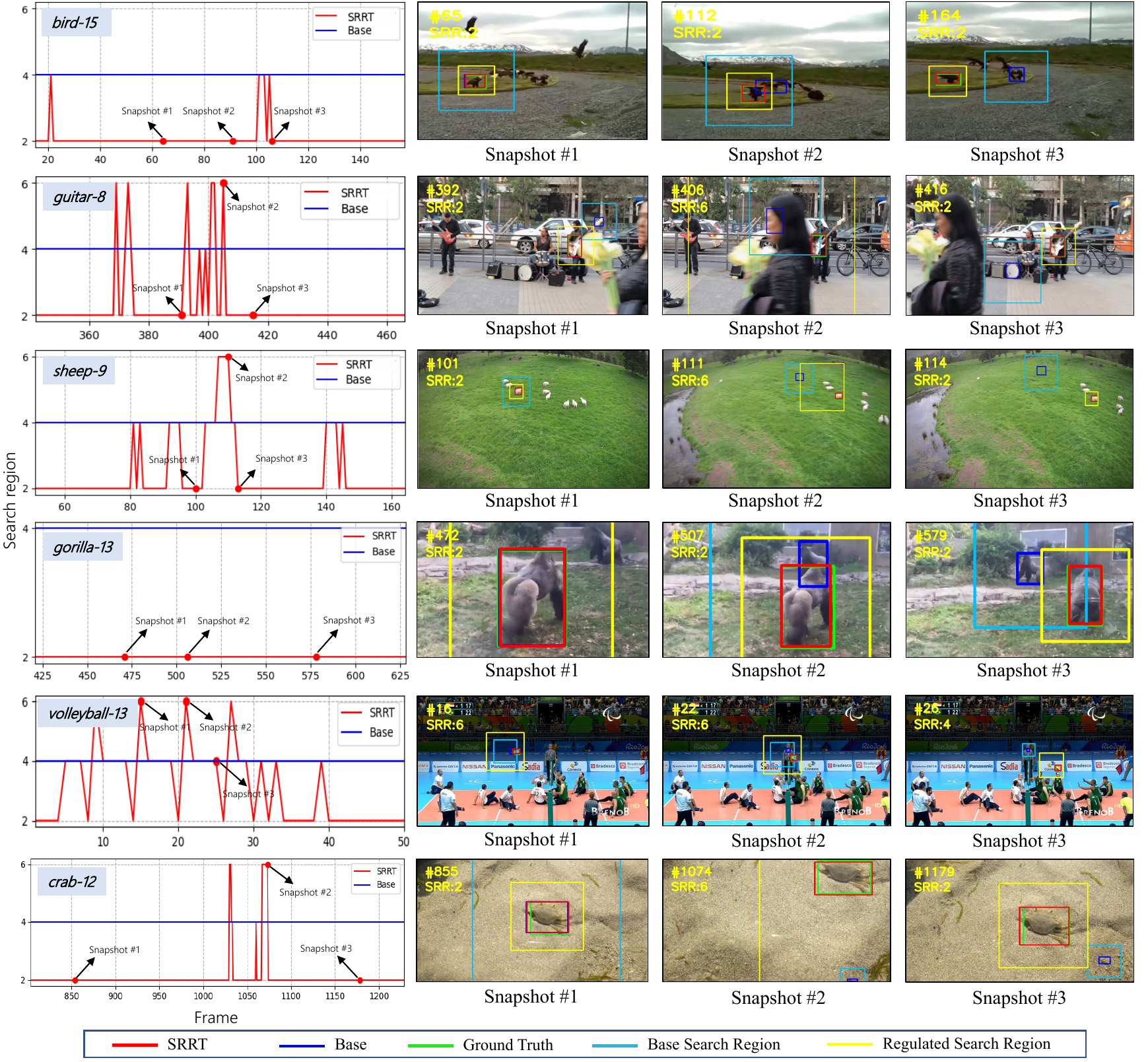}
	\caption{Representative cases with our SRRT scheme. Left: search region adoption curve, horizontal coordinate is the time step and vertical coordinate is the search region size. Right: typical snapshots extracted from the video sequence. Best viewed in color with zoom-in.}
	\label{fig:vis_example}
\end{figure*}

{\noindent \textbf{Why SRRT Works?}}
To analyze how SRRT works, we visualize several representative tracking clips during online tracking.
As shown in Figure~\ref{fig:vis_example}, benefiting from dynamic search region switching, 
SRRT runs smoothly with the smallest search region most of the time, except when the search radius needs to be increased to prevent the target object from being lost.
SRRT applies a small search region in most scenarios (\eg, ~target object has slight movement) which improves efficiency and avoids distractor objects (\eg,~ the first and fourth rows shown in Figure~\ref{fig:vis_example}). 
SRRT applies a large search region in some scenarios (\eg,~losing the target object) which can relocate the target position more precisely when the target object is lost 
(\eg, the third and the fifth rows shown in Figure~\ref{fig:vis_example}).
SRRT strives for an ideal tracking paradigm, that is, the tracker uses a narrow search region for stably tracking most of the time; when the target object is likely to be lost, the tracker can widen the search radius to relocate the target object.

\begin{figure}
	\begin{center}
		\includegraphics[width=0.45\textwidth]{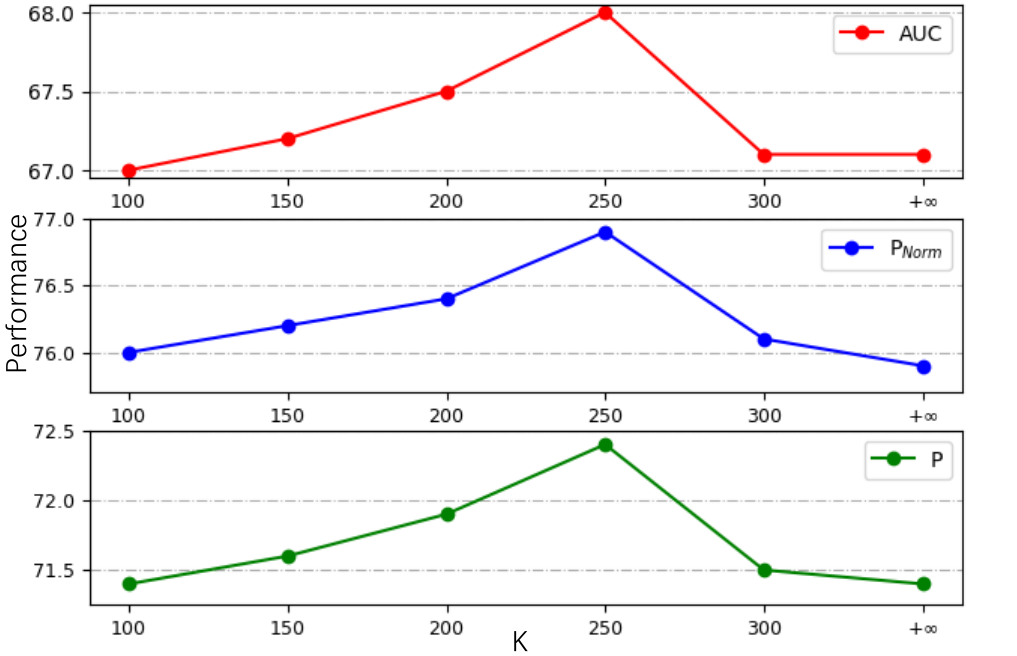}
		\caption{Performance variation against locking-state determined update thresholds $K$. AUC, P$_{Norm}$, and P scores on LaSOT benchmark are used as the metrics for threshold selection.} 
		\label{fig:ldu_plot}
	\end{center}
\end{figure}

{\noindent \textbf{Locking-state Determined Update.}}
As we describe in Section~\ref{sec:3.3}, a locking-state determined update strategy is proposed to update the dynamic reference frame for the search region regulator.
When the SRRT selects the smallest search region for consecutive $K$ frames, the model recognizes that the object is locked,
and updates the dynamic reference patch with the current frame. 
An accurate dynamic reference helps to deal with the challenge of object appearance changes in subsequent frames, providing more robust tracking performance.
Figure~\ref{fig:ldu_plot} shows the performance variation against locking-state 
threshold $K$. 
$K$=$+\infty$ represents never updating the dynamic reference frame.
As can be seen, SRRT reaches the best AUC, P$_{Norm}$, and P performance with the locking-state threshold of $K$=250. 
Too stringent or lax update conditions (higher or lower thresholds) will reduce the gains.
Compared to not updating 
\vspace{-0.5mm}
the dynamic reference frame, an appropriate update threshold 
can result in a 0.9\% performance gain in terms of AUC.

{\noindent \textbf{Different Size of Search Region.}}
To balance the search region requirements in different scenarios, existing trackers usually apply a mezzo fixed-size search region.
Therefore, the selected search region is inevitably large or small in some scenarios.
As shown in Tab.~\ref{tab:SR ablation}, $4SR$ obtains better performance than $2SR$, and a similar performance to $6SR$. 
When applying  a fixed search region, $4SR$ may be optimal. 
But as we discussed in Section~\ref{sec:introduction}, a small search region has the advantage of not being easily interfered with by other distractor objects, while a large search region has the ability to retrieve the lost target object. 
Therefore, is it possible to combine the advantages of small and large search regions by dynamic selection of them?
The answer is yes. We employ the proposed SRR, and we find the performance outperforms using any single search region.
$2SR/4SR$ surpasses both $2SR$ and $4SR$, and $4SR/6SR$ is preferred to both $4SR$ and $6SR$.
Finally, SRRT combines the advantages of each category of search region, delivering more advanced performance than any others.

\begin{table}[h]
	\centering
	\caption{Performance and speed comparison of different categories of search regions.}
	\begin{tabular}{@{\hspace{4pt}}c@{\hspace{4pt}}|@{\hspace{4pt}}c@{\hspace{4pt}}c@{\hspace{4pt}}@{\hspace{4pt}}c|@{\hspace{4pt}}c@{\hspace{3pt}}c@{\hspace{3pt}}}
		\toprule
		Method & AUC &~P$_{Norm}$ &P & Speed & Latency   \\
		\midrule
		$2SR$  &59.1 &66.7 &62.1   &51.9 &19.3ms  \\
		$2SR/4SR$ &65.1 &73.7 &69.2  &42.5 &23.5ms \\
		$4SR$ &64.9 &73.8 &69.0 &47.3 &21.1ms  \\
		$4SR/6SR$ &67.3 &76.6 &71.6 &36.8 &27.2ms \\
		$6SR$ &64.6 &74.1 &67.6 &26.9 &37.2ms  \\
		SRRT &68.0 &76.9 &72.4 &41.8 &23.9ms \\
		\bottomrule
	\end{tabular}
	\label{tab:SR ablation}
\end{table}

\begin{table}[h]
	\centering
	\caption{ Comparison of SRRT with different backbones.}
	\setlength{\tabcolsep}{0.9mm}{
		\begin{tabular}{c|ccc|cc}
			\toprule
			Method & AUC & P$_{Norm}$  & P & Speed & Latency \\
			\midrule
			Base &64.9 &73.8 &69.0 &47.3 &21.1ms \\
			\midrule
			+ SRR(Conv-5) &66.3 &75.5 &70.7 &49.3 &20.3ms \\
			+ SRR(ResNet-18) &67.1 &76.1 &71.4 &47.2 &21.2ms \\
			+ SRR(ResNet-34) &67.5 &76.5 &71.8 &44.6 &22.4ms \\
			+ SRR(ResNet-50) &68.0 &76.9 &72.4 &41.8 &23.9ms \\
			\bottomrule
	\end{tabular}}
	\label{tab:backbone_SRRTranst}
\end{table}

\begin{figure}[h]
	\centering
	\newcommand{\wid}{1.0\columnwidth}
	\includegraphics[width=0.47\textwidth]{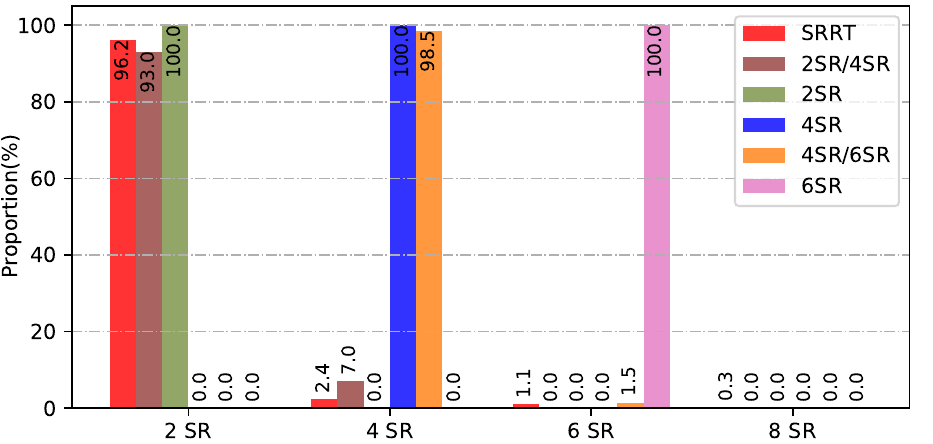}
	\caption{Distribution of regulated search regions. Under our SRRT paradigm, the smallest search region (2 $SR$) is adopted at most of the frames.}
	\label{fig:RSR_dis}
\end{figure}

{\noindent \textbf{Distribution of Regulated Search Region.}}
As we discuss in Section~\ref{sec:introduction}, a tracker can sufficiently capture the target object 
with a small size search region most time and Figure~\ref{fig:intro_statis} shows the minimum search region size distribution of adjacent frames on four benchmarks. 
Here, we calculate the regulated search region distribution under our SRRT scheme. 
The statistical results are shown in Figure~\ref{fig:RSR_dis}, and the corresponding performance results can be seen in Tab.~\ref{tab:SR ablation}.
Consistent with our motivation, SRRT runs with the smallest search region most of the time, which occupies 96.2\% of all the frames.
After a cliff-like descent, SRRT takes 2.4\%, 1.1\%, and 0.3\% of $4SR$, $6SR$, and $8SR$, respectively.
It indicates that SRRT only needs a small number of large search regions to prevent the target object from leaving the ROI.
Other search region regulation schemes ($2SR/4SR$ and $4SR/6SR$) also choose the smallest search region most of the time, which is in line with previous results.
It is worth noting that all regulation schemes outperform those using a fixed-size search region.

\begin{figure*}[t]
	\centering
	\includegraphics[width=1.0\textwidth,height=5cm]{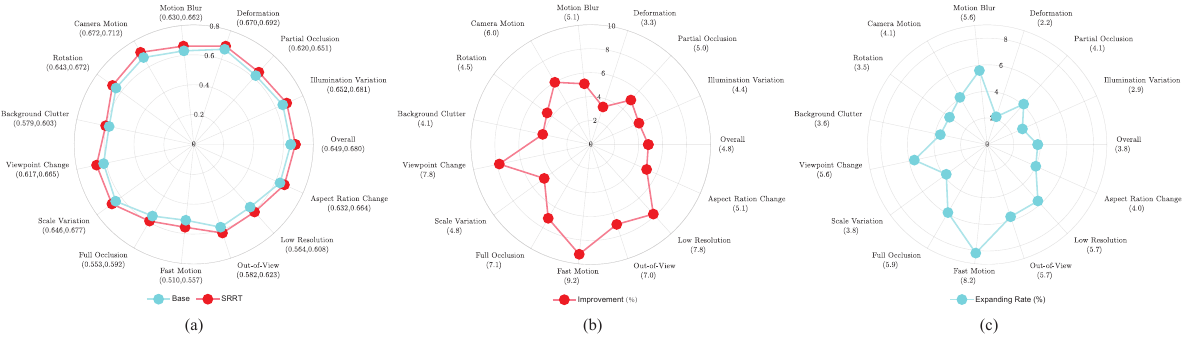}
	\vspace{-0.6cm}
	\caption{Attribute-wise analysis of SRRT. (a): SRRT brings an all-around improvement in all attributes. (b)\&(c): Several attributes with the greatest improvement are \textit{Fast Motion} and \textit{Viewpoint Change}, etc., which are also the scenarios where the fixed search region is prone to fail, and the results of the search region expansion rate also prove that SRRT switches the search region in these scenarios. Best viewed in color with zoom-in.}
	\label{fig:attr}
\end{figure*}

{\noindent \textbf{Attribute-wise Analysis.}}
Figure~\ref{fig:attr} shows the attribute-wise analysis of SRRT.
As shown in Figure~\ref{fig:attr} (a), the proposed SRRT outperforms the baseline method on all attributes (including challenging attributes of \textit{Full Occlusion}, \textit{Fast Motion}, and \textit{Low Resolution}).
Figure~\ref{fig:attr} (b) presents the improvement rate of SRRT on all 14 different attributes.
We find that the most significant improvements are from the sequences with the attribute of \textit{Fast Motion} (9.2\% relative improvement). Sequences with the attributes of 
\textit{Viewpoint Change} (7.8\%), \textit{Out-of-View} (7.0\%), and \textit{Full Occlusion} (7.1\%) also gain a large improvement. 
The results clearly demonstrate the advantages of SRRT's ability to rediscover the target objects and adaptability in the face of complex and extreme object motion states.
Another attribute that obtains large improvement is \textit{Low Resolution}.
The reason for the improvement on this attribute may be that SRRT chooses to use the smallest search region 94.3\% of the time for online tracking, which helps to obtain finer localization and bounding box estimation than using a large fixed-size search region.
Figure~\ref{fig:attr} (c) shows the search radius expanding rate\footnote{Expanding rate: the proportion of employing a search region larger than $2^2$ times the target object area.} on different attributes of challenge.
As can be seen, sequences with attributes that are prone to miss the target objects have a higher frequency to use a large search radius. 
In addition, since many challenge attributes are usually mixed (that is a video may include multiple challenges, \eg, fast motion and background clutter or more) in a video, sequences with other attributes are also evaluated with a corresponding boost, ultimately resulting in a performance boost for SRRT on almost all attributes.

{\noindent \textbf{Different Backbones.}}
As shown in Tab.~\ref{tab:backbone_SRRTranst}, we investigate the SRR module with different backbones. The ResNet-50 version  
achieves the best performance with deeper network layers and still has a speed of 41.8 $fps$. When employing a ResNet-18 backbone, the running speed (47.2 $fps$)  almost catches up with the speed of the original TransT (47.3 $fps$), yet 3.4\% relatively higher than that in terms of AUC.
SRRT has a slight impact on speed because it selects a small search region for the main tracking scenes.
Moreover, when we directly use a backbone of 5 convolutional layers (the first 5 layers in ResNet-18), surprisingly, the AUC also raises to 66.3\%, with a faster speed (49.3 $fps$) than the baseline; these results clearly demonstrate the validity of the SRRT paradigm.

{\noindent \textbf{Speed Influence.}}
We study the speed influence of different search regions and SRRT paradigm in Tab.~\ref{tab:SR ablation}. Experiments are run on an Nvidia RTX 2080Ti GPU.
With the $2^2$ times, $4^2$ times, and $6^2$ times search region, the base tracker 
has a speed of 51.9$fps$, 47.3$fps$, and 
26.9$fps$, respectively. The latency begins to increase when employing a large search region ($6SR$). $2SR/4SR$ and $4SR/6SR$ means that the search region is restricted to $2SR/4SR$ and $4SR/6SR$, respectively, running under the SRRT paradigm.
SRRT finally runs at a real-time speed 
and outperforms all other solutions.

{\noindent \textbf{SR-dedicated Trackers $vs$ a Single Base Tracker.}}
As we described in Section~\ref{sec:3.4}, we employ SR-dedicated trackers to handle search regions with different search radii. 
What about using a single tracker instead of SR-dedicated trackers?
To explore different SRRT schemes, we trained a Single base Tracker (STracker)  with different search region samples to learn to adapt to scenarios with different search region size distributions. We replace SR-dedicated trackers with STracker. The experimental results are shown in Tab.~\ref{tab:ST}. 
As can be seen, with $4SR$, STracker gains a better performance than the baseline method, this may be attributed to that training with samples from different search regions plays a similar effect to data augmentation. 
Due to the limited fitting ability of a single model, the performance drops significantly on other search regions ($2SR$ and $6SR$). 
After applying our search region regulator (SRR+STracker), the AUC score boosts to 66.1\%, which is a 0.7\% improvement compared to the best performance of using a single search region. 
Even with the improvement, there is still a gap (1.9\% in terms of AUC) with our SRRT scheme which employs SR-dedicated trackers.

\begin{table}
	\centering
	\renewcommand\arraystretch{1.0}
	\fontsize{8.5pt}{4mm}\selectfont
	\caption{ Performance comparison of SR-dedicated trackers and single base trackers. SR-dedicated trackers are shown to be superior to a single base tracker.}
	\setlength{\tabcolsep}{0.9mm}{
		\begin{tabular}{@{\hspace{10pt}}c@{\hspace{10pt}}|@{\hspace{10pt}}c@{\hspace{10pt}}c@{\hspace{10pt}}c@{\hspace{10pt}}}
			\toprule
			Scheme & AUC &P$_{Norm}$ &P  \\
			\midrule
			Base($4SR$)  &64.9 &73.8 &69.0  \\
			STracker($2SR$)  &56.4 &64.0 &59.7  \\
			STracker($4SR$) &65.4 &74.2 &69.0 \\
			STracker($6SR$) &47.8 &56.3 &38.9 \\
			SRR+STracker &66.1 &75.1 &70.4  \\
			SRRT (ours) &68.0 &76.9 &72.4  \\
			\bottomrule
		\end{tabular}
	}
	
	\label{tab:ST}
\end{table}

\noindent\textbf{Effectiveness on more tracking architectures.}
To further confirm the plug-and-play characteristics of the proposed tracking paradigm, we have added more recent trackers as our base trackers.
As shown in Table~\ref{tab:SRRSOTA}, the selected base trackers cover current mainstream tracking frameworks, including \textit{CNN model} (SiamRPN++~\cite{siamrpnplusplus}), \textit{CNN + Transformer model} (TransT~\cite{transt} and STARK~\cite{stark}), and \textit{Transformer model} (OSTrack~\cite{ostrack}, MixFormer~\cite{mixformer}, TATrack~\cite{tatrack}, MAT~\cite{zhao2023representation}, and SeqTrack~\cite{seqtrack}).
We can see that the proposed plug-and-play SRRT paradigm delivers consistent performance gains. 
Meanwhile, the efficiency losses of trackers are kept within a small range, thereby safeguarding their utility.

\begin{table}[h]
	\centering
	\fontsize{9pt}{4mm}\selectfont
	\caption{Comparison results on LaSOT and effectiveness upon multiple existing methods.} 
	\setlength{\tabcolsep}{1.75mm}{
		\begin{tabular}{c|c|ccc|c}
			\toprule
			Method &Architecture &AUC &P$_{Norm}$ &P &Speed  \cr
			\midrule
			SiamRPN++ &CNN &52.3 &59.1 &52.4 &56.0  \\
			TransT  &CNN+Trans &64.9 &73.8 &69.0  &47.3\\
			STARK-S50  &CNN+Trans &65.4 &74.4 &68.5 &42.2 \\
			OSTrack-256 &Trans &69.1 &78.7 &75.2  &105.0 \\
			MixFormer-1k &Trans &67.9 &77.3 &73.9 &25.0\\
			TATrack-B  &Trans &69.4 &78.2 &74.1 &14.1 \\
			MAT &Trans &67.8 &77.3 &- &120 \\
			SeqTrack-B256 &Trans &69.9 &79.7 &76.3 &40.0 \\
			\hline 
			SRRSiamRPN++ &CNN &56.9 &64.0 &57.1 &47.2  \\
			SRRTransT  &CNN+Trans &68.0 &76.9 &72.4  &41.8 \\
			SRRSTARK  &CNN+Trans  &68.3 &77.3  &71.6  &37.9 \\
			SRROSTrack &Trans &70.2  &79.3   &75.4  &87.7\\
			SRRMixFormer &Trans &69.6 &78.4   &74.5  &23.8\\
			SRRTATrack &Trans &70.7  &79.2 &75.5 &13.8 \\
			SRRMAT &Trans &69.3  &78.0 &- &91.7 \\
			SRRSeqTrack &Trans &70.7 &80.4 &77.2 &36.4 \\
			\bottomrule
		\end{tabular}
	}
	\label{tab:SRRSOTA}
\end{table}

\begin{figure}[h]
	\centering
	\includegraphics[width=0.49\textwidth]{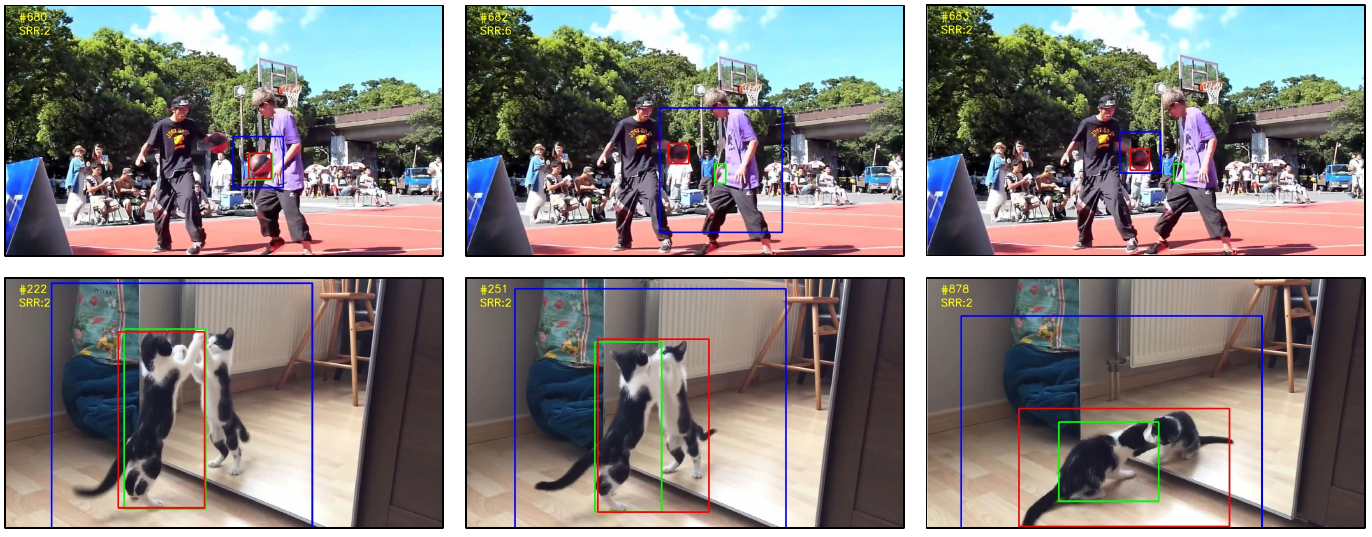}
	\vspace{-0.5cm}
	\caption{Failure cases of SRRT. The \textcolor{red}{red}, \textcolor{blue}{blue}, and \textcolor{green}{green} rectangular boxes represent the SRRT prediction results, the search region used in the current frame, and the ground truth, respectively.
	}
	\vspace{-0.3cm}
	\label{fig:fail_case}
\end{figure}

{\noindent \textbf{Failure cases.}}
We provide failure cases of the proposed SRRT as shown in Figure~\ref{fig:fail_case}. In the first case, SRRT lost the target object because the object was severely occluded or even almost disappeared, thus the tracking result was shifted to a nearby object and led to errors in subsequent frames. In addition, SRRT also fails to accurately identify the target object when facing challenges such as interference with similar targets ($i.e.$, the second case). SRRT can consider introducing additional multi-target trajectory modeling to cope with the case of occluded targets, and more powerful instance-level feature modeling to reduce the interference from similar objects.

\subsection{Limitation}
One limitation of SRRT is that, despite being more robust in dealing with some complex cases, SRRT separates the search region prediction and tracking processes rather than taking them as a whole, which can be more practical. 
Second, SRRT selects the optimal one from multiple fixed search regions, which can take advantage of the pre-trained tracker, while lacking some flexibility.
In future work, we are interested in further improving both the practicality and flexibility.

\section{Conclusion}

This paper proposes a novel tracking paradigm, search region regulation tracking (SRRT) for visual object tracking.
SRRT achieves accuracy and efficient tracking performance by applying a search region regulator for dynamically selecting an optimal search region size during online tracking. 
SRRT paradigm alleviates the difficulties in some intractable situations, \eg,~target object loss and distractor interference.
SRRT can be aggregated into existing trackers in a plug-and-play fashion incurring minimal additional lantency overhead,
significantly boosts the baseline trackers on eight widely used tracking benchmarks, and 
extensive experiments demonstrate its effectiveness.

\bibliographystyle{IEEEtran}
\bibliography{egbib}

\end{document}